\PassOptionsToPackage{table,dvipsnames}{xcolor}
\documentclass[sigconf,nonacm]{acmart}

\setcopyright{none}
\renewcommand\footnotetextcopyrightpermission[1]{}
\acmConference[KDD '27]{KDD 2027}{August 2027}{San Jose, CA, USA}
\acmYear{2027}
\acmDOI{}

\usepackage{booktabs}
\usepackage{graphicx}
\usepackage{subcaption}
\usepackage{amsmath}
\usepackage{multirow}

\usepackage{hyperref}
\usepackage{tikz}
\usetikzlibrary{decorations.pathreplacing, arrows.meta, fit, calc, positioning}

\usepackage{marvosym}
\usepackage{xcolor}
\usepackage{enumitem}
\usepackage{algorithm}
\usepackage{listings}

\makeatletter
\renewcommand{\@ACM@checkaffil}{}
\renewcommand{\@fnsymbol}[1]{\ifcase#1\or *\or\textdagger\or\textdaggerdbl\or\S\or\P\else\@ctrerr\fi}
\makeatother

\graphicspath{{figures/}}

\AtEndPreamble{%
  \hypersetup{%
    colorlinks=true,
    linkcolor=red,
    urlcolor=RoyalBlue,
    citecolor=cyan
  }%
}

\begin{document}

\title{FlexCast: Flexible Context Learning for Time Series Foundation Models}

\author{Zuyang Shen}
\authornote{Equal contribution.}
\affiliation{%
  \institution{Institute of Computing Technology, CAS}
}

\author{Luqi Gong}
\authornotemark[1]
\affiliation{%
  \institution{Zhejiang Lab}
}

\author{Fanda Fan}
\authornote{Corresponding author: Fanda Fan (\Letter: fanfanda@ict.ac.cn).}
\affiliation{%
  \institution{Institute of Computing Technology, CAS}
}

\author{Bingsong Wang}
\affiliation{%
  \institution{Beijing University of Posts and Telecommunications}
}

\author{Xiang Ao}
\affiliation{%
  \institution{Beijing Jiaotong University}
}

\author{Yahui Wei}
\affiliation{%
  \institution{Beijing Language and Culture University}
}

\author{Minghao Liu}
\affiliation{%
  \institution{Chengdu University of Technology}
}

\author{Chunjie Luo}
\author{Jianfeng Zhan}
\affiliation{%
  \institution{Institute of Computing Technology, CAS}
}

\begin{abstract}
\setlength{\emergencystretch}{1em}
Time series foundation models (TSFMs) based on autoregressive (AR) next-token
prediction entangle two factors: variable-context conditioning and
prediction-unit factorization with autoregressive rollout. Strong AR performance
therefore does not
reveal whether generalization stems from the flexible training distribution,
the local prediction target, the left-to-right inference process, or their
combination. We propose \textbf{FlexCast}, a flexible context-target learning
framework that disentangles these factors. FlexCast reformulates forecasting as
variable-length masked prediction over time series patches: during pretraining,
it dynamically samples visible context and masked target patches, exposing the
model to diverse context-target configurations without sequential decoding. At
inference time, the same pretrained model supports one-step forecasting,
confidence-based iterative refinement, and next-token-style generation through
a unified mask-and-predict interface, enabling controlled comparison of
inference schedules under identical architecture, data, and objective.
Evaluated on BLAST and multiple zero-shot benchmarks, FlexCast achieves
accuracy competitive with or better than strong TSFM baselines while
substantially reducing inference latency. Ablations confirm that context
diversity and contiguous target masking drive performance, while left-to-right
decoding does not yield consistent gains once flexible context-target
relationships have been learned---suggesting that the key benefit of AR-style
TSFMs lies in flexible training rather than sequential inference. The code is
available at \url{https://github.com/BenchCouncil/FlexCast}.
\end{abstract}



\maketitle

\section{Introduction}\label{sec:introduction}

\begin{figure}[tbp]
\centering
\includegraphics[width=\columnwidth]{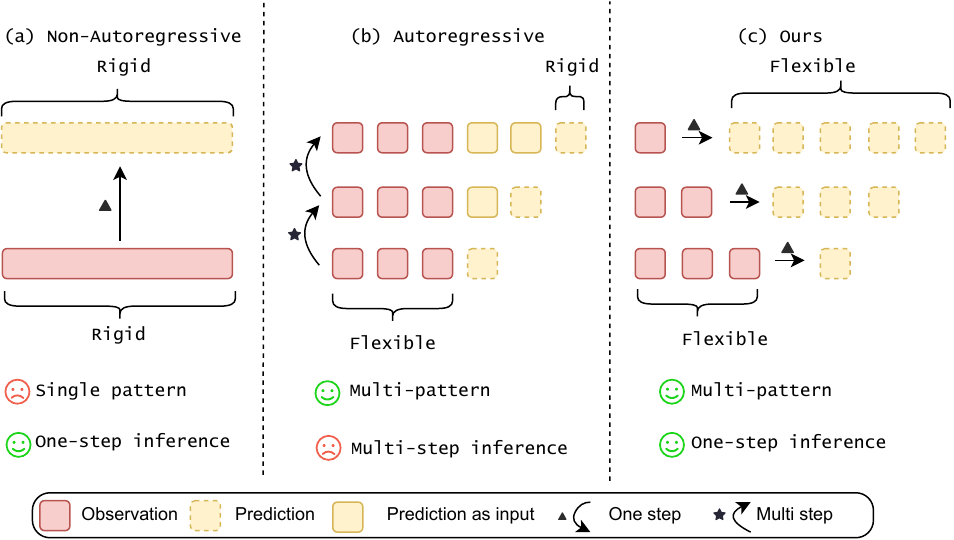}
\caption{Comparison of forecasting patterns and inference modes. \textbf{(a) Conventional Non-Autoregressive} forecasting is rigid: a task-specific model usually learns one fixed observation-length--prediction-length pattern, so new context or horizon settings require another model, head, or task configuration, although inference is one-step. Recent Non-AR TSFMs can relax this restriction through variable-length or masked training. \textbf{(b) Autoregressive} forecasting is flexible on the observation side: next-token training lets one model learn multiple variable-observation-to-unit-target distributions, but the full horizon must be generated by multi-step inference with predictions fed back as input. \textbf{(c) FlexCast} is flexible on both sides: masked forecasting lets one model learn multiple observation-length--prediction-length patterns while preserving one-step inference.}
\Description{A three-panel schematic comparing conventional Non-Autoregressive, Autoregressive, and FlexCast forecasting. The conventional Non-Autoregressive panel shows rigid observation and prediction windows, single-pattern training, and one-step inference. The Autoregressive panel shows flexible observation histories, prediction-as-input tokens, multi-pattern training, and multi-step inference. The FlexCast panel shows flexible observations and predictions with multi-pattern training and one-step inference.}
\label{fig:intro}
\end{figure}

Time series forecasting underpins cloud resource provisioning and business operations~\cite{huang2022dgraph,pan2023magicscaler,januschowski2024flexible}, traffic forecasting~\cite{guo2014towards,hu2018risk,hu2019stochastic,guo2020context,shao2022decoupled,lin2023origin}, and large-scale monitoring~\cite{pelkonen2015gorilla,adams2020monarch,shen2023lindorm,wang2023apache}. Traditional pipelines train separate models for each dataset, domain, sampling frequency, and horizon~\cite{qiu2024tfb,wang2026deep}, limiting reuse and raising maintenance costs as data change~\cite{ma2026positioninevitableendonearchitecturefitsalldomains}. Time series foundation models (TSFMs)~\cite{rasul2023lagllama,das2024decoder,ansari2024chronos,woo2024moirai,goswami2024moment,ansari2025chronos2,shi2025time,liu2025timerxl,liu2025sundial,wen2026revisitingtransformer,khwaja2026toto20timeseries} instead pretrain on heterogeneous corpora~\cite{shi2025time,aksu2024giftevalbenchmarkgeneraltime,shao2025blast} and transfer to unseen tasks zero- or few-shot, enabling universal forecasting across domains.

Yet it remains unclear what makes recent TSFMs generalize so well. A prominent line of TSFMs adopts the autoregressive (AR) paradigm, training models for context-conditioned prediction and generating forecasts through sequential decoding~\cite{liang2024foundation,ansari2024chronos,shi2025time}. This design is often treated as a single modeling choice. We instead ask a more diagnostic question: \emph{which part of the AR paradigm is actually responsible for the success of TSFMs?} We attempt to answer this question by separating AR forecasting into two coupled factors: the distribution of contexts seen by the model,\footnote{Throughout this paper, \emph{context} denotes the visible observations available to the model, while a \emph{prediction unit} denotes the future time steps or patches produced in one model call.} and the factorization of the forecast horizon into prediction units that are rolled out autoregressively. Figure~\ref{fig:intro} illustrates these factors through rigid or flexible observation-prediction patterns and single-pass or multi-step inference.

\textbf{(i)} Contexts are variable. Conventional task-specific Non-AR forecasters usually learn one mapping from a fixed-length context to a fixed-length horizon~\cite{Zeng2022AreTE,nie2023a,liu2024itransformer,qiu2025duet,hu2025timefilter}. Recent Non-AR TSFMs relax this restriction through variable-length or masked training~\cite{woo2024moirai,ansari2025chronos2,wen2026revisitingtransformer}, but they instantiate context flexibility through different training patterns. In contrast, an AR model obtains multiple effective contexts directly from next-token prediction: as the generation position changes, the model conditions on histories with different lengths and compositions. This \textbf{variable-context conditioning} may help TSFMs adapt to the diverse context lengths and forecasting configurations encountered in foundation-model evaluation, but its contribution remains entangled with sequential rollout.

\textbf{(ii)} Prediction is factorized into autoregressive units. A Non-AR forecaster typically predicts the entire horizon as a structured output in a single model call. In contrast, an AR forecaster partitions the horizon into smaller prediction units and generates them sequentially. Next-token prediction (NTP) is the special case in which each unit contains one token, whereas blockwise AR forecasting predicts $k$ future steps or patches in each call before feeding them back as context. This factorization couples prediction granularity with autoregressive rollout: smaller units may simplify each local prediction~\cite{huang2025timebase} and allow earlier forecasts to condition later ones~\cite{nagarajan2025roll}, but they also require repeated model calls and expose later predictions to accumulated errors. It is therefore unclear whether unit-wise AR rollout is necessary, or whether a model trained with flexible contexts can directly produce the full horizon as a structured prediction.

The difficulty is that flexible-context conditioning and unit-wise autoregressive rollout are coupled by the AR mechanism itself. AR forecasting creates changing contexts by repeatedly appending each predicted unit to the observed sequence. Consequently, strong AR performance does not reveal whether generalization comes from the diverse context distribution seen during training, from decomposing the horizon into smaller prediction units and rolling them out sequentially, or from their combination. To make this distinction testable, we need a framework that can learn from flexible contexts while independently controlling the prediction-unit size and inference schedule.

To test this distinction, we propose \textbf{FlexCast}, which reformulates forecasting as variable-length imputation over time series patches. During pretraining, it dynamically samples which patches are visible as context and which patches should be predicted. Varying the position and length of each contiguous masked span creates contexts with diverse lengths and compositions under the same objective, preserving the learning-side context diversity associated with AR forecasting without using sequential decoding to produce it.

At inference time, the same mask-and-predict interface supports single-pass and
multi-step forecasting. Quantile prediction provides patch-level
uncertainty~\cite{ansari2025chronos2,auer2026tirex,liu2026timers1billionscaletimeseries,wen2026revisitingtransformer},
allowing FlexCast to reveal high-confidence patches and recompute the remainder
under an expanded context in the spirit of MaskGIT~\cite{chang2022maskgit}.
The reveal schedule can be confidence-ordered or left-to-right, and revealed
patches remain fixed. Because all modes share the same pretrained model,
architecture, data, and objective, their comparison isolates prediction-unit
factorization and autoregressive rollout.

Following the balanced BLAST protocol~\cite{shao2025blast}, our experiments show
that flexible context learning provides the dominant and more consistent
benefit, whereas autoregressive rollout offers no consistent gain. FlexCast
achieves competitive zero-shot accuracy while \textbf{substantially reducing
the latency} of sequential decoding, and iterative generation helps mainly in
selected long-horizon or high-uncertainty settings. These results accord with
evidence that simple patch Transformers can be strong TSFMs when data, scaling,
and objectives are controlled~\cite{wen2026revisitingtransformer}.

This paper makes the following contributions:
\begin{itemize}[topsep=0.1cm, left=0.1cm]
    \item We decompose AR-based TSFMs into flexible-context conditioning and prediction-unit factorization with autoregressive rollout, turning their success into a testable attribution question.
    \item We propose \textbf{FlexCast}, a unified variable-length forecasting framework based on masked prediction. By varying the position and length of contiguous masked spans, FlexCast learns from flexible contexts with diverse lengths and compositions without relying on sequential decoding to create them.
    \item We design a unified inference interface for FlexCast that supports single-pass full-horizon forecasting and multi-step generation with controllable prediction units and reveal orders, including left-to-right autoregressive rollout and confidence-ordered progressive generation enabled by quantile uncertainty.
    \item Controlled experiments show that flexible context learning dominates the evaluated FlexCast factors, while autoregressive rollout is unnecessary for strong zero-shot performance. FlexCast remains competitive while reducing sequential-decoding latency.
\end{itemize}

\section{Related Work}\label{sec:related_work}

\subsection{Time Series Foundation Models}
Time series foundation models (TSFMs) pretrain large Transformer models on
heterogeneous corpora and transfer them to unseen forecasting tasks. Existing
TSFMs can be broadly grouped by their generation paradigm. Autoregressive TSFMs
such as TimesFM~\cite{das2024decoder}, Lag-Llama~\cite{rasul2023lagllama},
Chronos~\cite{ansari2024chronos}, Timer-XL~\cite{liu2025timerxl},
Time-MoE~\cite{shi2025time}, and Sundial~\cite{liu2025sundial} predict future
values through token-wise or blockwise autoregressive rollout. One source of
their
flexibility is the changing context distribution: at each generation step the model conditions on
a different effective context length, exposing it to diverse context
configurations during pretraining. However, sequential decoding at inference
time increases latency and may accumulate errors over long horizons.
Non-autoregressive TSFMs such as MOIRAI~\cite{woo2024moirai},
MOMENT~\cite{goswami2024moment}, Chronos-2~\cite{ansari2025chronos2}, and
PatchTST-FM~\cite{wen2026revisitingtransformer} produce the entire horizon in
one or few forward passes, avoiding sequential decoding.

Conventional task-specific Non-AR forecasters typically learn a fixed
context-to-horizon mapping. Recent Non-AR TSFMs relax this restriction in
different ways: MOIRAI samples forecasting configurations, Chronos-2 supports
variable-length prediction, and PatchTST-FM learns from masked patch patterns.
Thus, context flexibility is not exclusive to AR models, but these systems do
not isolate its contribution from mask geometry and
the inference schedule. FlexCast provides a common mask-and-predict interface
for this analysis while retaining efficient single-pass forecasting. The same
pretrained model can also switch between single-pass, confidence-ordered, and
autoregressive inference, enabling a controlled comparison of inference
schedules.

\subsection{Iterative Parallel Generation}
Non-autoregressive generation can replace sequential decoding with iterative
refinement.
Mask-Predict~\cite{ghazvininejad2019MaskPredict} first demonstrated this
paradigm for machine translation: a conditional masked language model generates
all target tokens in parallel, then iteratively re-masks and regenerates the
least-confident tokens over a fixed number of rounds. This approach achieves
near-autoregressive quality while decoding significantly faster.
MaskGIT~\cite{chang2022maskgit} extends masked iterative generation to image
synthesis. A bidirectional transformer is trained to predict randomly masked
visual tokens; at inference, it starts from a fully masked canvas and
progressively reveals tokens in order of prediction confidence following a
cosine schedule, requiring only 8--16 steps to produce high-fidelity images.
Muse~\cite{chang2023muse} scales this paradigm to text-conditioned image
generation at 900M parameters, confirming that parallel iterative decoding
remains efficient and high-quality at scale.

FlexCast draws on this iterative generation paradigm for time series
forecasting. Rather than using a fixed cosine schedule as in MaskGIT, FlexCast
leverages its probabilistic quantile output to derive a data-dependent
confidence measure (quantile interval width) for ordering the reveal of
predicted patches. This connects the masked generation framework to uncertainty
estimation in forecasting: patches with narrow predicted intervals are revealed
first, and the remaining masked patches are recomputed in subsequent rounds
under the expanded context. Revealed patches remain fixed. The same pretrained
model also supports single-pass and autoregressive inference, enabling
controlled comparison across generation schedules.

\subsection{Masked Modeling for Time Series}
Masked modeling has been widely adopted as a self-supervised pretraining
objective for time series. PatchTST~\cite{nie2023a} randomly masks a subset of
input patches and reconstructs them through a Transformer encoder, using the
learned representations for downstream forecasting after fine-tuning.
TimeMAE~\cite{cheng2026timemae} decouples visible and masked patches into an
asymmetric encoder-decoder structure and targets classification and regression
tasks after adaptation. Using multiple masked views,
SimMTM~\cite{dong2023simmtm} reconstructs masked time points and uses the
pretrained representations to initialize forecasting models.
TimesBERT~\cite{zhang2025timesbert} scales
BERT-style masked patch modeling to 260 billion time points and jointly trains
a functional token prediction task to capture inter-variate and domain-level
semantics, achieving strong performance on classification, imputation, and
anomaly detection.

These approaches share a common design: masking serves as a pretext task to
learn transferable representations, and downstream tasks still require
task-specific fine-tuning with a new prediction head. Their masking
configurations are also typically static---for example, TimesBERT uses a fixed
25\% random mask ratio with a fixed context length of 512. In contrast,
FlexCast treats variable-length masked prediction as the \emph{forecasting
objective itself} rather than an intermediate representation-learning step.
The sequence length, mask ratio, span length, and target position are randomly
sampled during pretraining, covering diverse forecasting and reconstruction
configurations. FlexCast
therefore performs zero-shot forecasting without fine-tuning, and its
mask-and-predict interface unifies single-pass, iterative, and autoregressive
inference under a single trained model.

\section{Method}\label{sec:method}

\begin{figure*}[tbp]
\centering
\includegraphics[width=\textwidth]{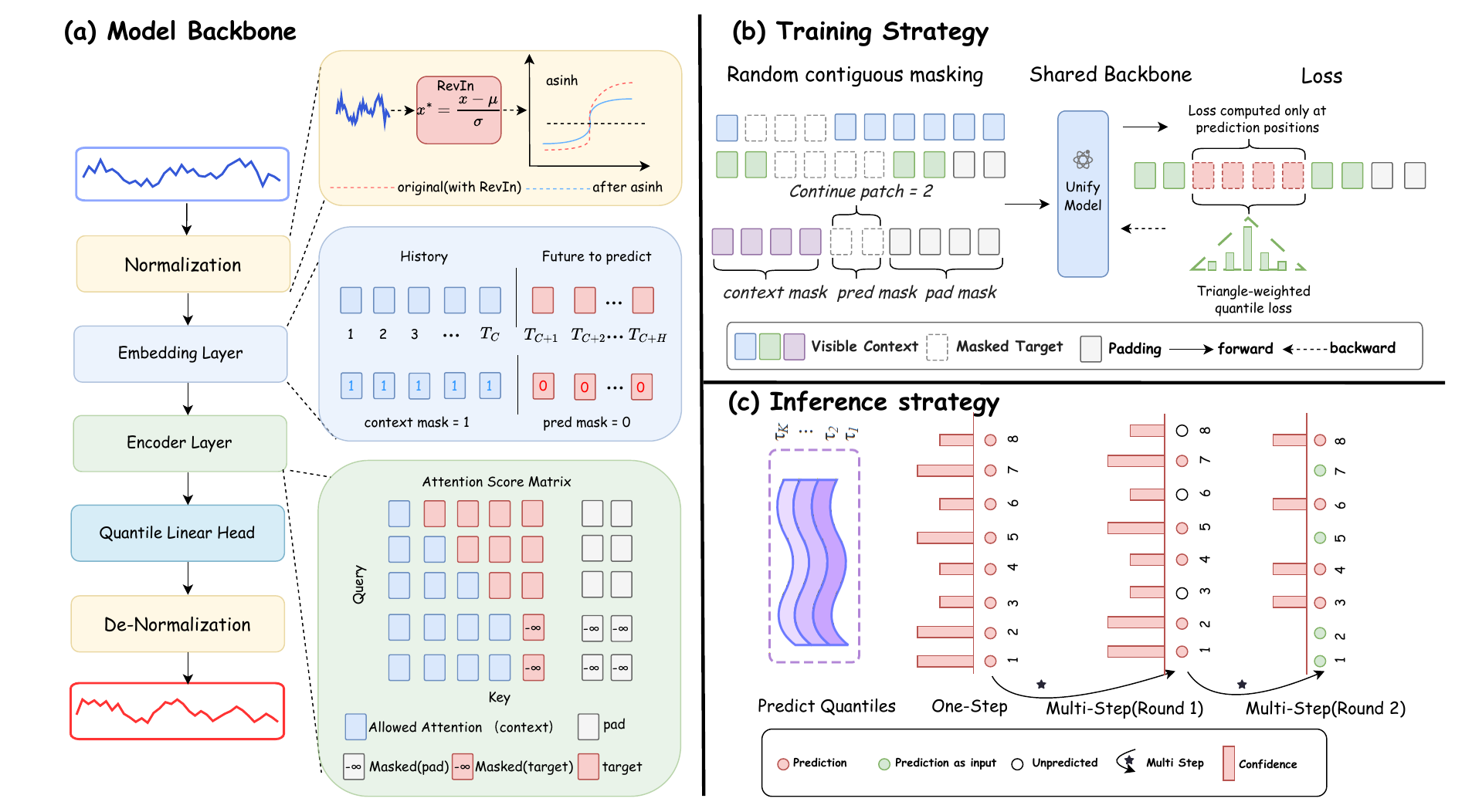}
\caption{Overview of the FlexCast framework. \textbf{(a) Model Backbone:} the input series is normalized with RevIN and an asinh transform, embedded together with context/prediction mask indicators, processed by a shared Transformer encoder whose additive attention bias sets padding keys to $-\infty$, and mapped to quantile forecasts before de-normalization. \textbf{(b) Training Strategy:} FlexCast uses random contiguous masking to sample visible context, masked target, and padding positions, feeds them into the shared backbone, and computes the quantile pinball loss only at prediction positions. \textbf{(c) Inference Strategy:} the same quantile predictor supports one-step forecasting and multi-step generation, where predicted quantiles provide confidence scores and selected predictions can be fed back as input in later rounds.}
\Description{Three-panel diagram of FlexCast. Panel a shows normalization, embedding with context and prediction mask indicators, a Transformer encoder with negative-infinity attention bias for padding keys, a quantile head, and de-normalization. Panel b shows random contiguous masking, a shared backbone, and quantile pinball loss at prediction positions. Panel c shows quantile prediction for one-step and multi-step inference, with confidence scores and predictions reused as input.}
\label{fig:method}
\end{figure*}

\subsection{Rethinking AR-based Forecasting}

Recent TSFMs widely adopt autoregressive (AR) generation from large sequence
models. Let $\mathbf{x}_{1:L}$ denote the observed historical sequence, where
$\mathbf{x}_t\in\mathbb{R}^{C}$ contains $C$ variables, and
$\mathbf{x}_{L+1:L+H}$ its forecast horizon, which is partitioned into
consecutive prediction units $\mathbf{u}_{1:M}$. An AR forecaster factorizes
the horizon as:
\begin{equation}
    p(\mathbf{x}_{L+1:L+H} \mid \mathbf{x}_{1:L})
    =
    \prod_{m=1}^{M}
    p(\mathbf{u}_m \mid \mathbf{x}_{1:L}, \mathbf{u}_{<m}).
\end{equation}
This formulation is adopted by TimesFM~\cite{das2024decoder},
Lag-Llama~\cite{rasul2023lagllama}, and
Chronos~\cite{ansari2024chronos}. In NTP, each $\mathbf{u}_m$ contains one
token; blockwise AR instead groups $k$ future steps or patches into each unit.
This factorization changes both the conditioning context and the prediction
granularity compared with a fixed-window non-AR mapping
$p(\mathbf{x}_{L+1:L+H} \mid \mathbf{x}_{1:L})$. At unit~$m$, the model
conditions on the original observations and all previously generated units,
so the effective context changes throughout the rollout. Smaller units may
simplify each local prediction, but producing the full horizon requires more
sequential model calls and feeds earlier errors into later contexts.

We therefore view AR-based forecasting as coupling two factors:
\emph{flexible-context conditioning} and \emph{prediction-unit factorization
with autoregressive rollout}. FlexCast makes their effects testable by
constructing variable-length imputation tasks in which the observed context and
prediction target are dynamically sampled, rather than relying on left-to-right
generation to create flexible contexts. The same pretrained model then supports
single-pass forecasting, confidence-ordered iterative generation, and
sequential decoding through one mask-and-predict interface
(Figure~\ref{fig:method}).

\subsection{Framework Overview}

FlexCast formulates forecasting as variable-length masked imputation. Because
masks specify both the context and target regions, one model can express any
context-to-horizon configuration without architectural changes.

FlexCast processes multivariate series channel-independently. For an input
$\mathbf{X}\in\mathbb{R}^{L\times C}$, each channel
$\mathbf{X}_{:,c}$ is treated as a separate univariate sequence, and the
channel axis is folded into the batch dimension. All channels share the same
patch embedding, Transformer encoder, and prediction head, but no attention is
computed across channels. In the following, we suppress the channel index for
clarity.

Given a univariate time series segment, we divide it into non-overlapping
patches $\mathbf{z}_{1:T}$ of length~$P$. The total number of patches~$T$ is
dynamically determined per instance during pretraining. Context patches retain
their original values; prediction target patches are zero-filled. A binary
mask indicator is concatenated to each patch so the model can distinguish
observed positions from positions to be inferred. The model processes the
full sequence of $T$ patches and outputs predictions for all masked positions
simultaneously.

Formally, FlexCast operates with three mutually exclusive patch-level masks:
\begin{equation}\label{eq:masks}
    \mathbf{m}^{\mathrm{ctx}},\; \mathbf{m}^{\mathrm{pred}},\; \mathbf{m}^{\mathrm{pad}} \in \{0,1\}^{T},
    \quad
    \mathbf{m}^{\mathrm{ctx}} + \mathbf{m}^{\mathrm{pred}} + \mathbf{m}^{\mathrm{pad}} = \mathbf{1}.
\end{equation}
Here $\mathbf{m}^{\mathrm{ctx}}_i{=}1$ marks patch~$i$ as visible context,
$\mathbf{m}^{\mathrm{pred}}_i{=}1$ marks it as a prediction target, and
$\mathbf{m}^{\mathrm{pad}}_i{=}1$ marks tail padding due to variable
sequence lengths. Context patches condition the model, prediction patches are
supervised by the loss, and padding patches are excluded from attention.
Varying these masks yields standard forecasting (target at tail),
imputation (target in middle), and AR next-token prediction (single-patch
target) as special cases of the same interface.

During training, FlexCast randomizes both the mask ratio and the span
position to expose the model to diverse context configurations.
During inference, the quantile-based probabilistic output enables a
confidence-ordered reveal schedule: rather than the fixed left-to-right order
of AR methods, FlexCast can reveal high-confidence patches first and recompute
the remaining masked patches under the expanded context. Revealed patches stay
fixed. When single-pass prediction suffices, all target patches are predicted
in a single forward pass.

Complete pseudocode appears as Algorithm~\ref{alg:flexcast} in
Appendix~\ref{app:method-details}.

\subsection{Backbone}

The FlexCast backbone consists of four modules: instance normalization, patch
embedding, a Transformer encoder, and a quantile prediction head.

\noindent
\textbf{Design rationale.}
FlexCast deliberately adopts a standard backbone to isolate the effect of
flexible context learning. Existing masked forecasters provide useful but
different masking designs. MOIRAI~\cite{woo2024moirai} samples context and
prediction lengths while retaining a tail-aligned forecasting layout. TiRex
introduces Contiguous Patch Masking (CPM)~\cite{auer2026tirex}, which distributes
targets across multiple contiguous blocks, and PatchTST-FM further shows that
such blocks outperform scattered masking~\cite{wen2026revisitingtransformer}.
Our ablation extends these findings: a single contiguous span with randomly
sampled position and length performs better than block-based CPM in our setting.
This design preserves a coherent forecasting target while exposing the model to
more diverse context configurations. Patching provides a practical balance
between local temporal modeling and computational efficiency. An encoder is
sufficient because the mask layout is known in advance and all target positions
can be predicted in parallel. The quantile head provides probabilistic forecasts
and interval-based confidence, allowing us to evaluate confidence-ordered
generation without an additional confidence estimator. Finally, we use
RoPE~\cite{su2023roformerenhancedtransformerrotary} to represent temporal order
and relative offsets under varying context layouts. In preliminary experiments,
training without RoPE was substantially less stable and often failed to
converge.

\noindent
\textbf{Instance normalization.}
Before patching, FlexCast applies reversible instance normalization
(RevIN)~\cite{kim2022reversible} per channel. The mean and standard deviation
are computed only from context-visible positions (positions where
$\mathbf{m}^{\mathrm{ctx}}{=}1$), excluding prediction targets and padding.
The normalized series is further stabilized by the $\mathrm{asinh}(\cdot)$
transform to compress heavy-tailed values. At output time, the inverse
transforms $\mathrm{sinh}(\cdot)$ and RevIN denormalization recover the
original scale.

\noindent
\textbf{Patch embedding.}
Following patch-based forecasting
models~\cite{nie2023a,liu2025timerxl,wen2026revisitingtransformer}, the
normalized series is segmented into non-overlapping patches of length~$P$.
Each patch $\mathbf{z}_i \in \mathbb{R}^{P}$ is paired with a binary mask
indicator $\bar{\mathbf{m}}_i \in \{0,1\}^{P}$ that marks observed time steps
($\bar{\mathbf{m}}_i = \mathbf{1}$ for context patches and
$\bar{\mathbf{m}}_i = \mathbf{0}$ for prediction patches). The two vectors are
concatenated and projected into a $d$-dimensional token via a residual block:
\begin{equation}
    \mathbf{e}_i =
    \mathrm{ResBlock}_{\mathrm{in}}\bigl([\mathbf{z}_i;\, \bar{\mathbf{m}}_i]\bigr),
\end{equation}
where $[\cdot;\cdot]$ denotes feature-axis concatenation.
$\mathrm{ResBlock}_{\mathrm{in}}$ consists of two linear projections with a
GELU activation and a skip connection:
$\mathrm{ResBlock}(\mathbf{u}) = \mathrm{Linear}_2(\mathrm{GELU}(\mathrm{Linear}_1(\mathbf{u}))) + \mathrm{Linear}_{\mathrm{skip}}(\mathbf{u})$,
projecting the $(P{+}P)$-dimensional input to $d$ dimensions. Zero-filling
$\mathbf{z}_i$ and setting $\bar{\mathbf{m}}_i{=}\mathbf{0}$ for prediction
patches explicitly signal that the model must infer these positions.

\noindent
\textbf{Transformer encoder.}
The embedded sequence is processed by a standard pre-norm Transformer
encoder~\cite{NIPS2017_3f5ee243} with rotary positional encoding
(RoPE)~\cite{su2023roformerenhancedtransformerrotary,shi2025time} applied
inside attention layers:
\begin{equation}
    \mathbf{h}_{1:T} =
    \mathrm{Encoder}(\mathbf{e}_{1:T},\, \mathbf{A}),
\end{equation}
where $\mathbf{A} \in \mathbb{R}^{T \times T}$ is an additive attention bias
that sets positions corresponding to padding patches
($\mathbf{m}^{\mathrm{pad}}_j{=}1$) to $-\infty$. All non-padding patches attend
to each other bidirectionally, regardless of whether they are context or
prediction patches.

\noindent
\textbf{Quantile prediction head.}
For probabilistic forecasting, let
$\mathcal{Q}=\{q_1<\cdots<q_Q\}$ denote the ordered quantile levels, where
$Q=|\mathcal{Q}|$. FlexCast predicts all $Q$ quantiles for each patch. A
residual output block maps each hidden state to a base prediction and a set of
non-negative offsets:
\begin{equation}
    \bigl[\hat{\mathbf{b}}_i,\;
    \hat{\boldsymbol{\delta}}_i^{(1)}, \ldots,
    \hat{\boldsymbol{\delta}}_i^{(Q)}\bigr]
    = \mathrm{ResBlock}_{\mathrm{out}}(\mathbf{h}_i),
\end{equation}
where $\hat{\mathbf{b}}_i \in \mathbb{R}^P$ and each
$\hat{\boldsymbol{\delta}}_i^{(k)} \in \mathbb{R}^P$. Quantile predictions
are constructed via cumulative summation:
\begin{equation}\label{eq:quantile}
    \hat{\mathbf{z}}_i^{(q_k)}
    = \hat{\mathbf{b}}_i
    + \sum_{j=1}^{k}
    \frac{\mathrm{softplus}\bigl(\hat{\boldsymbol{\delta}}_i^{(j)}\bigr)}{Q},
\end{equation}
which guarantees monotonicity across quantile levels by construction. When a
deterministic output is needed, FlexCast uses a triangularly weighted
combination that emphasizes central quantiles:
\begin{equation}\label{eq:point-forecast}
    \hat{\mathbf{z}}_i^{\mathrm{point}}
    = \sum_{q\in\mathcal{Q}} w_q\hat{\mathbf{z}}_i^{(q)},
    \qquad
    w_q =
    \frac{\frac{1}{2}-\left|q-\frac{1}{2}\right|}
    {\sum_{q'\in\mathcal{Q}}
    \left(\frac{1}{2}-\left|q'-\frac{1}{2}\right|\right)}.
\end{equation}

\subsection{Training Strategy}

FlexCast uses variable-length masked pretraining to construct diverse context
configurations analogous to those created by AR training through sequential
decoding. The training
pseudocode appears as Algorithm~\ref{alg:flexcast} in
Appendix~\ref{app:method-details}.

\noindent
\textbf{Dynamic effective length.}
Real-world time series corpora contain sequences of varying lengths. For each
training instance, FlexCast computes the effective length
$L_{\mathrm{eff}}$ --- the number of valid time steps excluding trailing
missing values or padding. Masks are constructed only within this valid
region, so the model never learns from padding artifacts. Because
$L_{\mathrm{eff}}$ varies across instances, the model naturally encounters
different total context lengths during training, analogous to the varying
prefix lengths in AR decoding but without sequential generation.

\noindent
\textbf{Contiguous span masking.}
Within the valid region, FlexCast samples a single contiguous span as the
prediction target. Unlike random token-level masks used in language
pretraining~\cite{devlin2019bert}, contiguous spans better match the structure
of forecasting targets, which are always temporally contiguous. The mask ratio
$r$ is drawn uniformly from a predefined range $[r_{\min}, r_{\max}]$,
yielding a span length $S = \lfloor L_{\mathrm{eff}} \cdot r / P \rfloor
\cdot P$ aligned to patch boundaries. The span start position is sampled
uniformly from $[0,\, L_{\mathrm{eff}} - S]$. This randomization exposes the
model to both extrapolation patterns (span at the end) and reconstruction
patterns (span in the interior), with varying ratios of context to target.

\noindent
\textbf{Loss function.}
FlexCast uses the quantile pinball loss
$\rho_q$~\cite{wen2026revisitingtransformer}, aggregated over every prediction
patch and quantile level:
\begin{equation}\label{eq:loss}
    \mathcal{L}
    =
    \frac{1}{\sum_{i}\mathbf{m}^{\mathrm{pred}}_i}
    \sum_{i=1}^{T}
    \mathbf{m}^{\mathrm{pred}}_i
    \cdot
    \frac{1}{|\mathcal{Q}|}
    \sum_{q\in\mathcal{Q}}
    \bigl\|
    \rho_q\bigl(\mathbf{z}_i,\, \hat{\mathbf{z}}_{i}^{(q)}\bigr)
    \bigr\|_1.
\end{equation}
The loss is computed in normalized $\mathrm{asinh}$ space; its element-wise
definition and normalization details appear in Appendix~\ref{app:training-loss}.

\subsection{Inference Strategy}

At inference time, FlexCast reuses the same mask-and-predict interface.
Different inference strategies correspond to different \emph{reveal
schedules} --- the order in which masked patches are predicted and converted
into context. Inference pseudocode appears in Algorithm~\ref{alg:flexcast}.

\noindent
\textbf{Single-pass full-horizon forecasting (one-step inference).}
The simplest strategy masks the entire forecast horizon and predicts all
target patches in a single forward pass. This is the most efficient mode
($\mathcal{O}(1)$ model calls) and serves as the primary evaluation setting:
strong single-pass performance directly demonstrates that flexible context
learning is sufficient without progressive generation.

\noindent
\textbf{Confidence-ordered iterative generation.}
Following the parallel decoding idea of
MaskGIT~\cite{chang2022maskgit}, FlexCast supports iterative generation over
$K$ rounds. At round~$k$, the model predicts all currently masked patches and
estimates per-patch confidence from the predicted quantile interval width:
\begin{equation}
    \mathrm{conf}_i
    =
    -\frac{1}{P}
    \left\|
    \hat{\mathbf{z}}_i^{(q_{\max})}
    -
    \hat{\mathbf{z}}_i^{(q_{\min})}
    \right\|_1,
\end{equation}
where a narrower interval indicates higher confidence. A fraction of the
most confident patches are revealed as context for round $k{+}1$. Specifically,
if the initial horizon contains $N_{\mathrm{pred}}$ target patches, each of the
first $K{-}1$ rounds reveals
$\lceil N_{\mathrm{pred}}/K\rceil$ of the currently masked patches (or all
remaining patches if fewer are available), and the final round predicts the
remainder. The reveal order is therefore determined by model uncertainty
rather than temporal order. Single-pass forecasting is the special case
$K{=}1$.

\noindent
\textbf{Autoregressive prediction-unit forecasting.}
FlexCast can also emulate AR inference as a special reveal schedule. At each
call, the next prediction unit is inferred and then revealed as context for
the subsequent call. A unit may contain one token, recovering NTP, or a block
of $k$ future steps or patches. Unlike confidence ordering, this schedule
always selects the leftmost remaining target unit and proceeds chronologically
until the horizon is complete. This mode uses the same pretrained model,
enabling direct comparison of unit-wise AR rollout against parallel and
confidence-ordered inference under a unified architecture and training
objective.

\section{Experiments}\label{sec:experiments}

We evaluate five questions that separate zero-shot accuracy and efficiency from
the effects of training and inference choices:
\begin{itemize}[topsep=0.1cm, left=0.1cm]
    \item \textbf{RQ1} (Section~\ref{sec:exp-zeroshot}): Can single-pass FlexCast deliver competitive zero-shot accuracy while reducing the latency of sequential TSFMs?
    \item \textbf{RQ2} (Section~\ref{sec:exp-ablation}): Which properties of flexible context learning are responsible for FlexCast's performance?
    \item \textbf{RQ3} (Section~\ref{sec:exp-inference}): After flexible-context learning, do multi-call unit-wise AR or confidence-ordered schedules improve over single-pass prediction?
    \item \textbf{RQ4} (Section~\ref{sec:exp-scaling}): How does FlexCast perform as model capacity increases within the evaluated range?
    \item \textbf{RQ5} (Section~\ref{sec:exp-sensitivity}): How sensitive is FlexCast to context length, tokenization, masking range, and quantile settings?
\end{itemize}

\subsection{Experimental Setup}\label{sec:exp-setup}

All models under our direct control are pretrained on the same BLAST
corpus~\cite{shao2025blast} and evaluated on the same held-out benchmarks.
Models available only as released weights, including Sundial~\cite{liu2025sundial}
and PatchTST-FM~\cite{wen2026revisitingtransformer}, provide additional
large-corpus reference points. We evaluate zero-shot forecasts on ETTh1, ETTh2,
ETTm1, ETTm2, Weather, and GlobalTemp~\cite{qiu2024tfb} at horizons 96, 192,
336, and 720, reporting MSE and MAE. FlexCast uses the point forecast in
Equation~\ref{eq:point-forecast} and single-pass inference unless stated
otherwise. Table~\ref{tab:experimental_settings} summarizes the default
configuration in Appendix~\ref{app:experimental-details}, together with the
corpus, baseline, hardware, and protocol details needed for reproduction.

The 4096-step context window is the maximum pretraining extent, not a fixed
observed-history budget. Sampling $r\sim U(0.1,0.3)$ partitions it into visible
context, one randomly positioned contiguous prediction span, and any required
padding. We use this moderate interval throughout the main experiments to vary
the effective context while keeping the prediction span learnable.

\subsection{Zero-shot Forecasting Performance}\label{sec:exp-zeroshot}

\textbf{Performance study.}
Table~\ref{tab:main_result} compares FlexCast with PatchTST-FM,
Sundial, and BLAST-retrained TSFMs~\cite{shao2025blast}. For fairness, FlexCast
and PatchTST-FM receive the longest BLAST evaluation context (3072), while
Sundial receives its supported maximum of 2880. FlexCast therefore receives no
extra history while composing a context and horizon combination not observed
during pretraining, and predicts each complete horizon in one call.

\begin{table*}[tbp]
  \definecolor{tableskyblue}{HTML}{63B7E6}
  \newcommand{\blastmark}{\ensuremath{^{\clubsuit}}}
  \newcommand{\patchmark}{\ensuremath{^{\blacklozenge}}}
  \newcommand{\sundialmark}{\ensuremath{^{\spadesuit}}}
  \caption{Zero-shot forecasting performance compared with strong forecasting baselines. The best results are marked in red, and the second-best results are marked in light blue. The symbols $s$, $b$, and $l$ represent the small, base, and large versions, respectively. Superscripts indicate different pretraining corpora: $\clubsuit$ for BLAST, $\blacklozenge$ for the PatchTST-FM corpus, and $\spadesuit$ for the Sundial corpus.}
  \label{tab:main_result}
  \resizebox{1\linewidth}{!}{
  \begin{tabular}{rr cc cc cc cc cc cc cc cc cc}
    \toprule
    \multicolumn{2}{l}{Models} & \multicolumn{2}{c}{FlexCast(Ours)\blastmark}
    & \multicolumn{2}{c}{PatchTST-FM\patchmark} & \multicolumn{2}{c}{Sundial$_{l}$\sundialmark} & \multicolumn{2}{c}{TimeMoE$_{l}$\blastmark} & \multicolumn{2}{c}{TimeMoE$_{b}$\blastmark} & \multicolumn{2}{c}{MOIRAI$_{l}$\blastmark} & \multicolumn{2}{c}{MOIRAI$_{b}$\blastmark} & \multicolumn{2}{c}{Chronos$_{b}$\blastmark} & \multicolumn{2}{c}{Chronos$_{s}$\blastmark} \\
    \multicolumn{2}{l}{\#Params} & \multicolumn{2}{c}{\textit{131.4M}}
    & \multicolumn{2}{c}{\textit{260M}} & \multicolumn{2}{c}{\textit{444M}} & \multicolumn{2}{c}{\textit{453M}} & \multicolumn{2}{c}{\textit{113M}} & \multicolumn{2}{c}{\textit{311M}} & \multicolumn{2}{c}{\textit{91M}} & \multicolumn{2}{c}{\textit{200M}} & \multicolumn{2}{c}{\textit{46M}} \\
    \cmidrule(r){3-4} \cmidrule(r){5-6} \cmidrule(r){7-8} \cmidrule(r){9-10} \cmidrule(r){11-12} \cmidrule(r){13-14} \cmidrule(r){15-16} \cmidrule(r){17-18} \cmidrule(r){19-20}
    \multicolumn{2}{l}{Metrics} & MSE & MAE & MSE & MAE & MSE & MAE & MSE & MAE & MSE & MAE & MSE & MAE & MSE & MAE & MSE & MAE & MSE & MAE \\
    \midrule
\multirow{5}{*}{\rotatebox{90}{ETTh1}} & 96 & .362 & .381 & .358 & .378 & \textcolor{red}{.346} & .383 & \textcolor{tableskyblue}{.348} & \textcolor{red}{.375} & .352 & \textcolor{tableskyblue}{.376} & .359 & .383 & .362 & .384 & .357 & \textcolor{red}{.375} & .359 & \textcolor{tableskyblue}{.376} \\
 & 192 & .395 & .403 & .400 & .407 & \textcolor{tableskyblue}{.386} & .410 & \textcolor{red}{.381} & \textcolor{red}{.399} & .389 & \textcolor{tableskyblue}{.401} & .395 & .404 & .400 & .405 & .397 & \textcolor{tableskyblue}{.401} & .403 & .402 \\
 & 336 & \textcolor{tableskyblue}{.409} & \textcolor{red}{.414} & .438 & .439 & .410 & .426 & \textcolor{tableskyblue}{.409} & .424 & \textcolor{red}{.408} & .419 & .411 & \textcolor{tableskyblue}{.416} & .416 & .420 & .422 & .417 & .431 & \textcolor{tableskyblue}{.416} \\
 & 720 & .461 & .458 & .500 & .464 & .438 & .459 & .447 & .451 & .450 & .455 & \textcolor{red}{.420} & \textcolor{red}{.430} & \textcolor{tableskyblue}{.430} & \textcolor{tableskyblue}{.439} & .460 & .443 & .449 & \textcolor{tableskyblue}{.439} \\
 & AVG & .407 & .414 & .424 & .422 & \textcolor{red}{.395} & .420 & \textcolor{tableskyblue}{.396} & .412 & .399 & .412 & \textcolor{tableskyblue}{.396} & \textcolor{red}{.408} & .402 & .412 & .409 & \textcolor{tableskyblue}{.409} & .410 & \textcolor{red}{.408} \\
    \midrule
\multirow{5}{*}{\rotatebox{90}{ETTh2}} & 96 & \textcolor{red}{.266} & \textcolor{red}{.320} & .293 & .342 & \textcolor{tableskyblue}{.269} & .330 & .276 & .329 & .285 & .332 & .288 & .325 & .284 & .324 & .282 & \textcolor{tableskyblue}{.321} & .281 & .326 \\
 & 192 & \textcolor{tableskyblue}{.332} & \textcolor{red}{.365} & .362 & .390 & \textcolor{red}{.325} & .373 & .345 & .376 & .348 & .378 & .353 & .370 & .348 & \textcolor{tableskyblue}{.369} & .356 & \textcolor{tableskyblue}{.369} & .353 & .371 \\
 & 336 & \textcolor{tableskyblue}{.360} & \textcolor{tableskyblue}{.386} & .408 & .431 & \textcolor{red}{.354} & .400 & .384 & .416 & .372 & .405 & .369 & \textcolor{red}{.382} & .367 & \textcolor{tableskyblue}{.386} & .378 & .397 & .387 & .403 \\
 & 720 & \textcolor{red}{.353} & \textcolor{red}{.394} & .419 & .462 & .389 & .443 & .442 & .470 & .419 & .452 & \textcolor{tableskyblue}{.387} & \textcolor{tableskyblue}{.406} & \textcolor{tableskyblue}{.387} & .410 & .403 & .424 & .411 & .430 \\
 & AVG & \textcolor{red}{.328} & \textcolor{red}{.366} & .371 & .406 & \textcolor{tableskyblue}{.334} & .387 & .361 & .397 & .356 & .391 & .349 & \textcolor{tableskyblue}{.370} & .346 & .372 & .355 & .377 & .358 & .382 \\
    \midrule
\multirow{5}{*}{\rotatebox{90}{ETTm1}} & 96 & .289 & \textcolor{tableskyblue}{.322} & \textcolor{tableskyblue}{.276} & \textcolor{red}{.312} & \textcolor{red}{.273} & .329 & .327 & .343 & .334 & .350 & .355 & .355 & .348 & .354 & .310 & .327 & .314 & .331 \\
 & 192 & .330 & \textcolor{tableskyblue}{.348} & \textcolor{tableskyblue}{.316} & \textcolor{red}{.333} & \textcolor{red}{.312} & .357 & .368 & .378 & .388 & .386 & .388 & .380 & .385 & .378 & .363 & .360 & .364 & .365 \\
 & 336 & .355 & \textcolor{tableskyblue}{.366} & \textcolor{tableskyblue}{.348} & \textcolor{red}{.359} & \textcolor{red}{.343} & .378 & .373 & .396 & .400 & .412 & .399 & .387 & .410 & .394 & .410 & .387 & .391 & .417 \\
 & 720 & \textcolor{tableskyblue}{.400} & \textcolor{tableskyblue}{.396} & \textcolor{tableskyblue}{.400} & \textcolor{red}{.391} & \textcolor{red}{.397} & .413 & .445 & .438 & .457 & .451 & .429 & .413 & .448 & .416 & .477 & .427 & .452 & .521 \\
 & AVG & .343 & \textcolor{tableskyblue}{.358} & \textcolor{tableskyblue}{.335} & \textcolor{red}{.349} & \textcolor{red}{.331} & .369 & .378 & .388 & .394 & .399 & .392 & .383 & .397 & .385 & .390 & .375 & .380 & .408 \\
    \midrule
\multirow{5}{*}{\rotatebox{90}{ETTm2}} & 96 & \textcolor{red}{.162} & \textcolor{red}{.242} & .175 & .260 & \textcolor{tableskyblue}{.172} & .255 & .180 & .259 & .181 & .260 & .192 & .259 & .194 & .265 & .175 & .249 & .180 & \textcolor{tableskyblue}{.248} \\
 & 192 & \textcolor{red}{.215} & \textcolor{red}{.280} & .229 & .294 & \textcolor{tableskyblue}{.227} & .296 & .245 & .305 & .247 & .307 & .256 & .302 & .257 & .304 & .242 & \textcolor{tableskyblue}{.290} & .243 & .292 \\
 & 336 & \textcolor{red}{.260} & \textcolor{red}{.313} & .282 & .333 & \textcolor{tableskyblue}{.275} & .331 & .283 & .338 & .293 & .344 & .289 & .329 & .301 & .342 & .299 & \textcolor{tableskyblue}{.326} & .302 & .331 \\
 & 720 & \textcolor{red}{.339} & \textcolor{red}{.367} & .347 & \textcolor{tableskyblue}{.374} & \textcolor{tableskyblue}{.343} & .378 & .364 & .392 & .376 & .396 & .372 & .384 & .387 & .396 & .394 & .387 & .406 & .396 \\
 & AVG & \textcolor{red}{.244} & \textcolor{red}{.301} & .258 & .315 & \textcolor{tableskyblue}{.254} & .315 & .268 & .323 & .274 & .326 & .277 & .318 & .284 & .326 & .277 & \textcolor{tableskyblue}{.313} & .282 & .316 \\
    \midrule
\multirow{5}{*}{\rotatebox{90}{Weather}} & 96 & \textcolor{red}{.152} & \textcolor{red}{.192} & \textcolor{tableskyblue}{.154} & .203 & .157 & .208 & .161 & .209 & .163 & .213 & .168 & .200 & .171 & .202 & .163 & \textcolor{tableskyblue}{.197} & .164 & .198 \\
 & 192 & \textcolor{red}{.196} & \textcolor{tableskyblue}{.235} & \textcolor{tableskyblue}{.202} & .252 & .207 & .256 & .217 & .261 & .215 & .263 & .246 & \textcolor{red}{.217} & .218 & .247 & .210 & .241 & .213 & .244 \\
 & 336 & \textcolor{red}{.252} & \textcolor{tableskyblue}{.277} & .265 & .306 & \textcolor{tableskyblue}{.259} & .295 & .276 & .304 & .273 & .297 & .288 & \textcolor{red}{.275} & .278 & .291 & .264 & .282 & .273 & .288 \\
 & 720 & .361 & .346 & \textcolor{red}{.325} & .341 & \textcolor{tableskyblue}{.327} & .342 & .342 & .353 & .328 & \textcolor{tableskyblue}{.339} & .351 & .375 & .370 & .350 & .339 & \textcolor{red}{.334} & .349 & .342 \\
 & AVG & .240 & \textcolor{red}{.262} & \textcolor{red}{.236} & .275 & \textcolor{tableskyblue}{.238} & .275 & .249 & .281 & .244 & .278 & .263 & .266 & .259 & .272 & .244 & \textcolor{tableskyblue}{.263} & .249 & .268 \\
    \midrule
\multirow{5}{*}{\rotatebox{90}{GlobalTemp}} & 96 & \textcolor{tableskyblue}{.216} & \textcolor{red}{.341} & \textcolor{red}{.211} & .344 & -- & -- & .226 & .346 & .229 & .349 & .234 & .351 & .236 & .352 & .226 & \textcolor{tableskyblue}{.343} & .230 & .345 \\
 & 192 & \textcolor{tableskyblue}{.260} & .384 & \textcolor{red}{.250} & \textcolor{tableskyblue}{.383} & -- & -- & .263 & .386 & .272 & .390 & .266 & \textcolor{red}{.382} & .268 & .384 & .271 & .384 & .280 & .389 \\
 & 336 & \textcolor{tableskyblue}{.296} & .420 & \textcolor{red}{.285} & \textcolor{red}{.416} & -- & -- & .309 & .420 & .311 & .423 & .309 & .420 & .309 & .420 & .314 & \textcolor{tableskyblue}{.419} & .318 & .420 \\
 & 720 & .360 & .472 & \textcolor{red}{.336} & .456 & -- & -- & \textcolor{tableskyblue}{.340} & \textcolor{red}{.447} & .343 & \textcolor{tableskyblue}{.449} & .361 & .459 & .347 & \textcolor{tableskyblue}{.449} & .427 & .502 & .438 & .504 \\
 & AVG & \textcolor{tableskyblue}{.283} & .404 & \textcolor{red}{.270} & \textcolor{tableskyblue}{.400} & -- & -- & .284 & \textcolor{red}{.399} & .288 & .402 & .292 & .403 & .290 & .401 & .322 & .418 & .316 & .414 \\
    \midrule
    \multicolumn{2}{c}{\# Wins} & \textcolor{red}{11} & \textcolor{red}{13} & 7 & 6 & 9 & 0 & 1 & 4 & 1 & 0 & 1 & 6 & 0 & 0 & 0 & 2 & 0 & 1 \\
    \bottomrule
  \end{tabular}}
\end{table*}

FlexCast achieves competitive zero-shot accuracy across datasets and horizons.
PatchTST-FM is pretrained on its own large corpus built from
GiftEvalPretrain, KernelSynth-style synthetic series, and TSMixup
data~\cite{wen2026revisitingtransformer}, while Sundial is pretrained on
TimeBench, a trillion-scale time-series corpus covering domains such as
finance, IoT, meteorology, and healthcare~\cite{liu2025sundial}. Although both
baselines use larger pretraining corpora than FlexCast, FlexCast remains
comparable to these large-corpus baselines and obtains the largest number of
best results in both MSE and MAE, with especially strong performance on ETTh2,
ETTm2, and Weather. Thus, learning across flexible contexts supports strong
zero-shot forecasting without sequential autoregressive decoding at inference
time.

Compared with the models retrained on the same BLAST corpus, FlexCast shows a
clearer advantage. It achieves the largest number of best results among the
BLAST-retrained TimeMoE, MOIRAI, and Chronos variants and improves over them on
most dataset-horizon settings, indicating that the gain does not come only from
the pretraining corpus but also from the proposed flexible mask-and-predict
training interface.

\begin{figure}[tbp]
  \centering
  \includegraphics[width=\columnwidth]{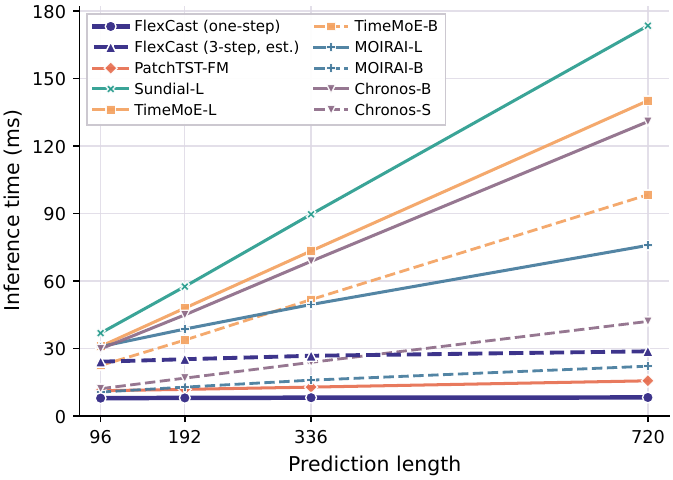}
  \caption{Inference-latency scaling across four prediction lengths with a fixed 512-step context. The three-step FlexCast curve and shorter-horizon baseline points are estimated from the measured endpoints. Lower is better.}
  \Description{Line chart comparing measured endpoint and estimated scaling trends for one-step and three-step FlexCast and baseline time series foundation models at prediction lengths 96, 192, 336, and 720.}
  \label{fig:latency_results}
\end{figure}

\noindent
\textbf{Inference efficiency study.}
Figure~\ref{fig:latency_results} compares latency scaling on one A100 GPU with
batch size 1, one channel, a fixed 512-step context, and four horizons. The
common 512-step context is supported by all implementations, whereas a
3376-step context is not. FlexCast predicts each horizon in one forward pass,
so its latency grows only moderately with horizon length. Sequential Sundial,
TimeMoE, and Chronos incur increasing rollout cost through repeated model calls,
and the latency gap therefore widens for longer horizons. FlexCast is also more
parameter-efficient than their larger variants.

FlexCast is also faster than the single-pass MOIRAI and PatchTST-FM baselines,
although parameter count alone does not explain these cases. MOIRAI uses
patch-size-specific input and output projections and maps its representations to
a mixture distribution, from which point forecasts are obtained by
sampling~\cite{woo2024moirai}. FlexCast instead uses one fixed patch size and
directly aggregates its predicted quantiles through
Equation~\ref{eq:point-forecast}. PatchTST-FM pads inputs to mark prediction and
padding positions under learned positional embeddings, whereas FlexCast uses
RoPE and excludes only actual tail padding from attention. These shorter input
and output paths are consistent with the measured latency gap, although the
study does not isolate their individual contributions.

\subsection{Flexible Context Learning Ablation}\label{sec:exp-ablation}

\textbf{Variable context.}
Following Section~\ref{sec:method}, we test whether the changing effective
contexts created by AR rollout, rather than the rollout itself, explain
generalization.

We compare four FlexCast variants that differ in whether the context and target
are rigid or flexible. As summarized in
Table~\ref{tab:diverse_pattern_settings}, the Rigid Context, Rigid Target
variant uses a single context-target geometry, where a 3376-step context
predicts the next 720 steps. The Flexible Context, Rigid Target variant samples
the context length from $\{3376, 2880, 1440\}$ while keeping the target length
fixed at 720. The Rigid Context, Flexible Target variant fixes the context at
3376 and varies the target length among $\{96, 192, 336, 720\}$. The full
Flexible Context, Flexible Target setting uses a 4096-step window and
interval-sampled mask rates to create both variable effective contexts and
variable target spans. To isolate these two factors, all four variants in this
ablation use tail-aligned contiguous masking; the effect of mask position is
examined separately in the subsequent masking-strategy ablation.
For a fair comparison, all variants remove samples whose effective sequence
length is shorter than 4096 before training and evaluation.

\begin{table}[tbp]
  \caption{Settings for the variable context-target ablation.}
  \label{tab:diverse_pattern_settings}
  \centering
  \small
  \setlength{\tabcolsep}{2pt}
  \definecolor{oursrowgreen}{HTML}{E8F5E9}
  \newcommand{\idsym}[1]{\makebox[1.1em][c]{\raisebox{-0.08ex}{#1}}}
  \newcommand{\stratlabel}[2]{\shortstack[l]{#1,\\[-0.15ex]#2}}
  \begin{tabular*}{\columnwidth}{@{}p{0.36\columnwidth}@{\extracolsep{\fill}}p{0.32\columnwidth}p{0.24\columnwidth}@{}}
    \toprule
    Strategy & Maximum context window ($X$) & Target ($Y$) \\
    \midrule
    \idsym{\textcolor[HTML]{668CB5}{\scalebox{0.95}{$\blacktriangle$}}} \stratlabel{Rigid Context}{Rigid Target} & 3376 & 720 \\
    \idsym{\textcolor[HTML]{86C2BE}{\rotatebox[origin=c]{45}{\rule{0.50em}{0.50em}}}} \stratlabel{Flexible Context}{Rigid Target} & 3376 / 2880 / 1440 & 720 \\
    \idsym{\textcolor[HTML]{F4C979}{\scalebox{0.78}{$\bigstar$}}} \stratlabel{Rigid Context}{Flexible Target} & 3376 & 96 / 192 / 336 / 720 \\
    \rowcolor{oursrowgreen}[0pt][0pt]
    \idsym{\textcolor[HTML]{CD686B}{\scalebox{0.95}{$\spadesuit$}}} \stratlabel{Flexible Context}{Flexible Target (Ours)} & 4096, random mask-rate & $r \sim U(0.1, 0.3)$ \\
    \bottomrule
  \end{tabular*}
\end{table}

\begin{figure}[tbp]
  \centering
  \includegraphics[width=\columnwidth]{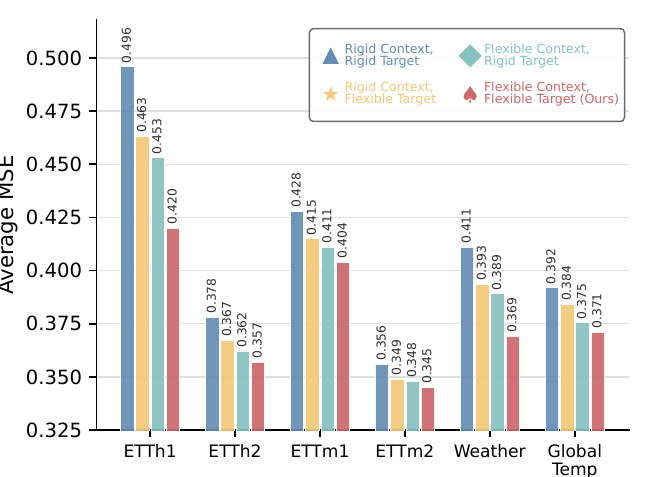}
  \caption{Average MSE of different context-target pattern designs across downstream datasets.}
  \Description{Bar chart comparing average MSE for rigid and flexible context-target pattern designs. The full flexible context and flexible target setting performs best.}
  \label{fig:diverse_pattern_results}
\end{figure}

Figure~\ref{fig:diverse_pattern_results} shows that context diversity is the
main source of improvement. The Rigid Context, Rigid Target setting gives the
weakest performance because it exposes the model to only one forecasting
geometry. Flexible Context with a rigid target already reduces the error
consistently, whereas Flexible Target with a rigid context is helpful but less
effective. The full Flexible Context, Flexible Target variant performs best
because it combines variable effective contexts and variable target lengths
under the same tail-aligned masking constraint used by the other variants.
Because every target is placed at the sequence tail, no variant can condition
on observations to the right of the target; the gain from flexible contexts
therefore persists under a forecasting-aligned, history-only information
pattern.

\noindent
\textbf{Masking strategy.}
We further study how the training objective and mask distribution affect the
forecasting problems seen during pretraining.
Table~\ref{tab:mask_strategy_settings} compares five strategies: Tail, which
always places the masked target at the end of the sequence; Contiguous Patch Masking
(CPM)~\cite{auer2026tirex},
which samples random short masked chunks scattered throughout the sequence;
Fixed-rate, which randomly places a contiguous masked segment but samples the
mask rate from $\{0.1,0.2,0.3\}$; and Random-rate, our default strategy,
which samples a continuous mask rate from $U(0.1,0.3)$ and masks one
contiguous segment at a random position. NTP is a FlexCast variant that uses
the same random mask-rate interval as Tail, always places the supervised span
at the sequence tail, and applies causal next-patch training followed by
left-to-right autoregressive rollout.

\begin{table}[tbp]
  \caption{Settings for the pretraining-strategy ablation.}
  \label{tab:mask_strategy_settings}
  \centering
  \small
  \setlength{\tabcolsep}{2pt}
  \definecolor{oursrowgreen}{HTML}{E8F5E9}
  \newcommand{\idsym}[1]{\makebox[1.1em][c]{\raisebox{-0.08ex}{#1}}}
  \begin{tabular*}{\columnwidth}{@{}p{0.25\columnwidth}@{\extracolsep{\fill}}p{0.32\columnwidth}p{0.12\columnwidth}p{0.21\columnwidth}@{}}
    \toprule
    Strategy & Mask rate & Position & Shape \\
    \midrule
    \idsym{\textcolor[HTML]{668CB5}{\scalebox{0.95}{$\clubsuit$}}} Tail & $r \sim U(0.1, 0.3)$ & Tail & Contig. \\
    \idsym{\textcolor[HTML]{86C2BE}{\scalebox{0.95}{$\spadesuit$}}} CPM & $r_{\max}=0.4,\ c_{\max}=6$ & Random & Contig. blocks \\
    \idsym{\textcolor[HTML]{F4C979}{\rotatebox[origin=c]{45}{\rule{0.50em}{0.50em}}}} Fixed & $r \in \{0.1, 0.2, 0.3\}$ & Random & Contig. \\
    \idsym{\textcolor[HTML]{9B83B4}{\scalebox{0.85}{$\blacktriangle$}}} NTP & $r \sim U(0.1, 0.3)$ & Tail & Next-patch \\
    \rowcolor{oursrowgreen}[0pt][0pt]
    \idsym{\textcolor[HTML]{CD686B}{\scalebox{0.95}{$\varheartsuit$}}} Random (Ours) & $r \sim U(0.1, 0.3)$ & Random & Contig. \\
    \bottomrule
  \end{tabular*}
\end{table}

\begin{figure}[tbp]
  \centering
  \includegraphics[width=\columnwidth]{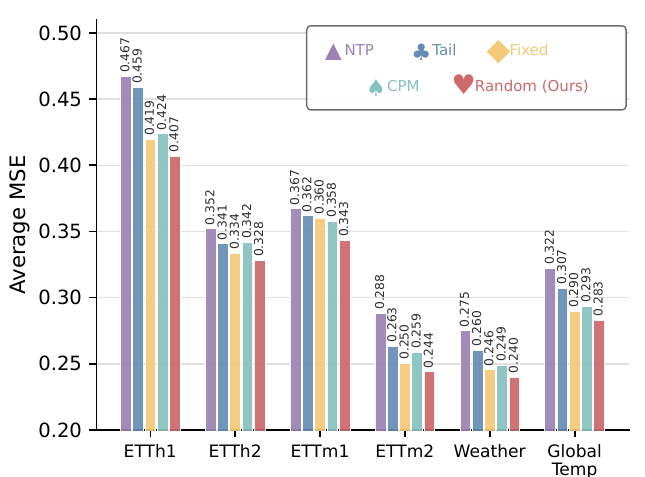}
  \caption{Average MSE of different pretraining strategies across downstream datasets.}
  \Description{Bar chart comparing Tail, Contiguous Patch Masking, Fixed-rate, NTP, and Random-rate pretraining strategies by average MSE. NTP has the highest error and Random-rate contiguous masking has the lowest error across the evaluated datasets.}
  \label{fig:mask_strategy_results}
\end{figure}

Figure~\ref{fig:mask_strategy_results} shows that Random-rate performs best,
whereas NTP trails even Tail on every dataset. Under the same backbone and
mask-rate interval, causal next-patch supervision at the sequence tail therefore
does not improve accuracy. Random target positions prevent specialization to
end-of-sequence prediction, while random rates expose the model to a continuum
of target lengths. A single contiguous span nevertheless remains closer to
forecasting than scattered local imputation and encourages the model to infer
longer-term temporal trends rather than relying mainly on neighboring values.
Together with the context ablation, these results show that context diversity is
the stronger factor, while flexible targets provide complementary prediction
granularities. Random-rate contiguous masking thus preserves useful learning-side
diversity without causal supervision or sequential rollout.

\subsection{Inference Strategy}\label{sec:exp-inference}

Using the same checkpoint and downstream datasets, we fix the context length to
3376 and evaluate horizons of 96, 192, 336, and 720 under three schedules:
(1)~one-step inference,
which predicts the complete horizon in a single forward pass;
(2)~confidence-based multi-step inference with $K{=}3$ rounds, which at each
of the first two rounds reveals the most confident one-third of the initial
target patches (measured by quantile interval width) and predicts all remaining
patches in the final round; and (3)~blockwise autoregressive inference, which
ignores confidence, predicts the leftmost remaining unit, and appends the
generated values to the visible context before the next call. Token-wise NTP
is the one-token special case; here we evaluate blockwise prediction units of
96 and 192 time steps.

\begin{figure}[tbp]
  \centering
  \includegraphics[width=\columnwidth]{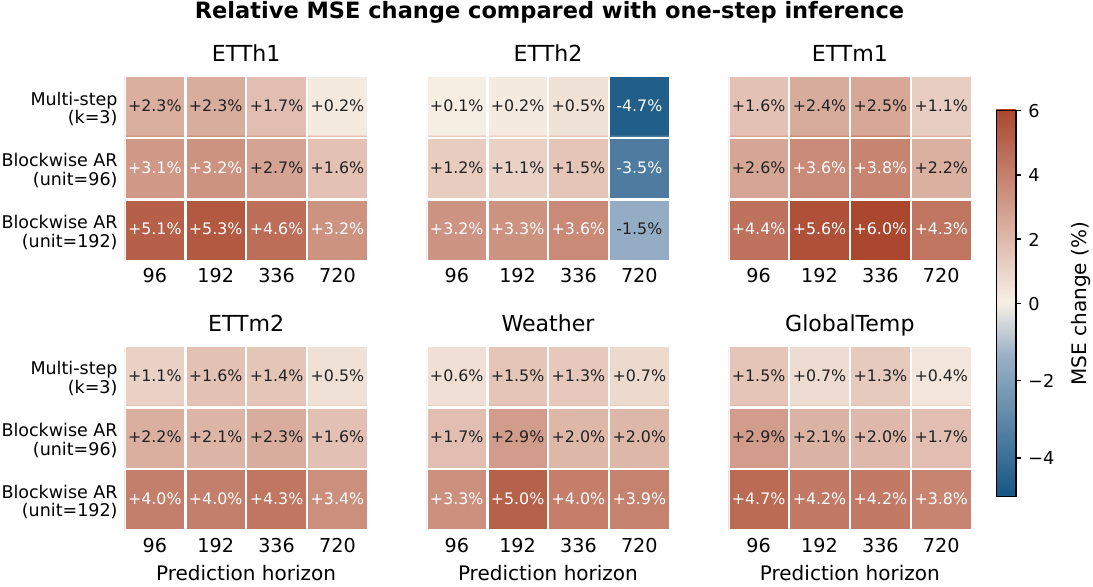}
  \caption{Relative MSE change compared with one-step inference across datasets, horizons, and inference schedules. Positive values indicate higher MSE than one-step inference, while negative values indicate lower MSE.}
  \Description{Relative MSE comparison of one-step, confidence-based multi-step, and blockwise autoregressive inference schedules across datasets and horizons.}
  \label{fig:inference_strategy_results}
\end{figure}

Figure~\ref{fig:inference_strategy_results} shows that confidence-based
multi-step inference remains close to one-step results in most settings and
improves ETTh2 at the 720-step horizon by 4.7\%. It can therefore help for long,
uncertain horizons without requiring left-to-right generation. By contrast,
blockwise AR usually raises MSE, and 192-step units are generally worse than
96-step units. Appending predictions to the context exposes later calls to
earlier errors, making the result sensitive to prediction granularity.
Single-pass inference is thus sufficient in most settings; when extra calls do
help, confidence ordering preserves the mask-and-predict interface without a
forced chronological order.

\subsection{Model Scaling}\label{sec:exp-scaling}

To study how capacity affects forecasting, we train FlexCast$_{s}$ (15M
parameters, 6 layers, $d{=}256$), FlexCast$_{b}$ (40M, 8 layers, $d{=}512$),
and FlexCast$_{l}$ (131.4M, 10 layers, $d{=}1024$). All variants share the same
BLAST pretraining data, training steps, mask-rate range, and evaluation protocol;
only model architecture differs. Full configurations appear in
Table~\ref{tab:scaling_law} in Appendix~\ref{app:exp-scaling-results}.

Average MSE improves monotonically from 0.360 for FlexCast$_{s}$ to 0.332 for
FlexCast$_{b}$ and 0.308 for FlexCast$_{l}$. Each capacity increase improves
performance within the evaluated range, and FlexCast$_{l}$ reduces average MSE
by approximately 14\% relative to FlexCast$_{s}$. This trend indicates that
additional model capacity benefits the flexible mask-and-predict objective under
the controlled training budget.

\subsection{Model Parameter Sensitivity Analysis}\label{sec:exp-sensitivity}

We evaluate context-length, architecture, and objective sensitivity. Under the
fixed BLAST budget, the same checkpoint performs best near 3376 observed steps
and the default settings remain stable; full results appear in
Appendix~\ref{app:exp-sensitivity-results}. Across the controlled sweeps, a
patch length of 16, the $[0.1,0.3]$ mask-rate interval, and 99 quantiles perform
best, while moving beyond these defaults provides no consistent benefit under the
same optimization budget.

\section{Conclusion}\label{sec:conclusion}

FlexCast separates flexible-context conditioning from autoregressive rollout
through variable-length masked prediction. Across BLAST and public benchmarks,
it achieves competitive zero-shot accuracy at lower latency. Ablations show that
flexible history and random-position contiguous masking improve accuracy,
whereas chronological rollout offers no consistent gain. Within the evaluated
settings, context diversity therefore captures the primary AR benefit, making
single-pass forecasting the efficient default.



\bibliographystyle{ACM-Reference-Format}
\bibliography{sample}

\clearpage
\appendix
\raggedbottom
\section{Additional Method Details}\label{app:method-details}

\subsection{Training-Loss Details}\label{app:training-loss}
For target $\mathbf{z}_i$ and prediction
$\hat{\mathbf{z}}_{i}^{(q)}$, the element-wise pinball loss is
\begin{equation}
    \rho_q\bigl(\mathbf{z}_i,\, \hat{\mathbf{z}}_{i}^{(q)}\bigr)
    = 2\left|
    \bigl(\mathbf{z}_i - \hat{\mathbf{z}}_{i}^{(q)}\bigr)
    \cdot
    \bigl(\mathbb{I}[\mathbf{z}_i \le \hat{\mathbf{z}}_{i}^{(q)}] - q\bigr)
    \right|.
\end{equation}
The objective in Equation~\ref{eq:loss} is computed in the normalized
$\mathrm{asinh}$ space so that large-magnitude series do not dominate the
gradient. Averaging over $|\mathcal{Q}|$ gives each predicted quantile equal
weight while keeping the loss scale comparable across different numbers of
quantile levels.


\definecolor{codeblue}{rgb}{0.25,0.5,0.5}
\definecolor{codesign}{RGB}{0, 0, 255}
\definecolor{codefunc}{rgb}{0.85, 0.18, 0.50}
\definecolor{codenetwork}{RGB}{106, 90, 205}
\lstdefinelanguage{FlexCastPython}{
  language=Python,
  alsoletter={_},
  keywordstyle=\color{blue}\bfseries,
  commentstyle=\color{codeblue},
  stringstyle=\color{orange},
  showstringspaces=false,
  literate=*
    {+}{{\color{codesign}+ }}{1}
    {-}{{\color{codesign}- }}{1}
    {*}{{\color{codesign}* }}{1}
    {/}{{\color{codesign}/ }}{1}
    {def}{{\color{codefunc}def }}{1}
    {return}{{\color{codefunc}return }}{1}
    {effective_length}{{\color{codefunc}effective\_length}}{16}
    {uniform}{{\color{codefunc}uniform}}{7}
    {align_to_patch}{{\color{codefunc}align\_to\_patch}}{14}
    {randint}{{\color{codefunc}randint}}{7}
    {build_span_mask}{{\color{codefunc}build\_span\_mask}}{15}
    {asinh}{{\color{codefunc}asinh}}{5}
    {sinh}{{\color{codefunc}sinh}}{4}
    {revin_normalize}{{\color{codefunc}revin\_normalize}}{15}
    {revin_inverse}{{\color{codefunc}revin\_inverse}}{13}
    {pinball}{{\color{codefunc}pinball}}{7}
    {concat}{{\color{codefunc}concat}}{6}
    {zeros}{{\color{codefunc}zeros}}{5}
    {ones}{{\color{codefunc}ones}}{4}
    {topk_confident}{{\color{codefunc}topk\_confident}}{14}
    {leftmost_masked}{{\color{codefunc}leftmost\_masked}}{15}
    {count_masked_patches}{{\color{codefunc}count\_masked\_patches}}{20}
    {ceil}{{\color{codefunc}ceil}}{4}
    {any}{{\color{codefunc}any}}{3}
    {triangular_point}{{\color{codefunc}triangular\_point}}{16}
    {model}{{\color{codenetwork}model}}{5}
}

\lstset{
  language=FlexCastPython,
  backgroundcolor=\color{white},
  basicstyle=\fontsize{7.5pt}{8.5pt}\ttfamily\selectfont,
  columns=fullflexible,
  breaklines=true,
  captionpos=b,
}

\begin{algorithm*}[tbp]
\centering
\caption{FlexCast training and inference.}
\label{alg:flexcast}

\begin{minipage}[t]{0.48\linewidth}
\noindent\textbf{Training.}\par
\noindent{\scriptsize Variable-length masked pretraining with dynamic span masking.}\par
\begin{minipage}{0.98\linewidth}
\begin{lstlisting}[language=FlexCastPython, escapechar=`]
# model: FlexCast encoder
# x: raw time series [B, T, C]
# channels are folded into the batch and processed independently
# P: patch length
# Q: number of quantile levels

# compute effective length (exclude tail NaN)
eff_len = effective_length(x)  # [B]
# identify tail padding for variable-length samples
pad_mask = positions(T) >= eff_len[:, None]  # [B,T]

# dynamic mask ratio per sample
r = uniform(r_min, r_max, size=B)
span_len = align_to_patch(eff_len * r, P)

# random contiguous span within valid region
start = randint(0, eff_len - span_len)
start = align_to_patch(start, P)
pred_mask = build_span_mask(start, span_len, T) & ~pad_mask
ctx_mask = ~(pred_mask | pad_mask)

# normalization (RevIN + asinh)
x_norm = asinh(revin_normalize(x, ctx_mask))

# forward: zero-fill targets and exclude padding from attention
x_input = x_norm * ctx_mask[..., None]
attn_bias = where(pad_mask, -inf, 0)
q_pred = model(x_input, pred_mask, attn_bias)  # [B,T,C,Q]

# pinball loss on masked region only
loss = pinball(q_pred, x_norm, pred_mask) / Q
loss.backward()
\end{lstlisting}
\end{minipage}
\end{minipage}
\hfill
\begin{minipage}[t]{0.48\linewidth}
\noindent\textbf{Inference.}\par
\noindent{\scriptsize Supports one-step, iterative, and AR modes.}\par
\begin{minipage}{0.98\linewidth}
\begin{lstlisting}[language=FlexCastPython, escapechar=`]
# x_ctx: observed context [B, L, C]
# H: forecast horizon (in time steps)
# P: patch length
# mode: single, confidence, or left_to_right
# K: confidence rounds (K=1: single-pass)
# U: prediction-unit size for left_to_right
# model: pretrained FlexCast encoder

# build initial sequence: context + masked horizon
x = concat(x_ctx, zeros(B, H, C))
mask = concat(zeros(B, L, C), ones(B, H, C))

# normalization
x_norm = asinh(revin_normalize(x, 1 - mask))

# initial number revealed per confidence round
n_target = count_masked_patches(mask, P)
n_reveal = ceil(n_target / K)
k = 0

while True:
    x_input = x_norm * (1 - mask)
    q_pred = model(x_input, mask)  # [B,T,C,Q]

    if mode == "single" or (mode == "confidence" and k == K - 1):
        break

    if mode == "confidence":
        # reveal N/K highest-confidence target patches
        width = q_pred[..., -1] - q_pred[..., 0]
        reveal = topk_confident(width, mask, n_reveal)
    else:  # chronological AR schedule
        reveal = leftmost_masked(mask, U)

    x_norm[reveal] = triangular_point(q_pred[reveal])
    mask[reveal] = 0
    k += 1
    if not any(mask):
        break

# compose the horizon: retain values revealed in prior rounds
# and fill only the patches that remain masked in the final round
pred_norm = x_norm.clone()
pred_norm[mask == 1] = triangular_point(q_pred[mask == 1])
forecast = revin_inverse(sinh(pred_norm[:, L:, :]))
\end{lstlisting}
\end{minipage}
\end{minipage}
\Description{Two pseudocode panels. The first describes FlexCast training with effective-length computation, dynamic contiguous span masking, normalization, model prediction, and masked pinball loss. The second describes inference with a masked horizon, confidence-ordered progressive reveal or left-to-right generation, and final de-normalization.}
\end{algorithm*}

\section{Detailed Experimental Setup}\label{app:experimental-details}

\noindent
\textbf{Unified evaluation protocol.}
Existing TSFMs are pretrained on different corpora of varying scale and
composition, making it difficult to attribute performance differences to model
design rather than data advantages. To control for this confound, we adopt a
unified experimental protocol: all models under our direct control are
pretrained on the same BLAST corpus~\cite{shao2025blast} and evaluated on the
same held-out benchmarks, so that observed differences reflect architectural
and objective choices rather than data scale. For industry models that only
release inference weights (e.g., Sundial~\cite{liu2025sundial},
PatchTST-FM~\cite{wen2026revisitingtransformer}), we include them in the
zero-shot comparison as additional reference points. Because their pretraining
corpora are typically larger than BLAST, this comparison is conservative for
FlexCast. All ablation experiments strictly follow this unified setting to
ensure comparable conclusions.

\noindent
\textbf{Pretraining corpus.}
FlexCast is pretrained on BLAST~\cite{shao2025blast}, a balanced time-series
corpus for universal forecasting. The official BLAST split contains 19.8
million training sequences and 0.2 million validation sequences, each with
4096 time steps. Since FlexCast uses a patch length of 16, each training
sequence is converted into 256 patch tokens, yielding approximately 5.07
billion training patch tokens in total. BLAST is constructed from large-scale
public time-series sources, including climate and weather datasets (CMIP6,
ERA5, WeatherBench), building-energy data (Buildings-900K), and hundreds of
other public datasets.

For scale context, BLAST contains approximately 81 billion time points from
balanced public sources, including CMIP6, ERA5, and Buildings-900K. The
PatchTST-FM corpus~\cite{wen2026revisitingtransformer} combines
GIFT-Eval-Pretrain (approximately 230 billion observations), KernelSynth (10
million synthetic series), and TSMixup (4 million augmented series). The
Sundial corpus (TimeBench)~\cite{liu2025sundial} contains approximately one
trillion time points from ERA5, LOTSA, the Chronos collection, ECG, finance,
and IoT sources.

\noindent
\textbf{Evaluation benchmarks.}
For zero-shot evaluation, we use six public
benchmarks~\cite{qiu2024tfb}: ETTh1, ETTh2, ETTm1, ETTm2, Weather, and
GlobalTemp. These datasets are the main public benchmarks used in the BLAST
evaluation and are widely adopted for zero-shot forecasting; they are excluded
from the BLAST pretraining corpus to avoid evaluation leakage. Following the
common long-term forecasting protocol, we evaluate four prediction horizons,
96, 192, 336, and 720, and report results without any target-dataset
fine-tuning.

\noindent
\textbf{Baselines and metrics.}
We compare FlexCast with PatchTST-FM~\cite{wen2026revisitingtransformer}, the
autoregressive Sundial model~\cite{liu2025sundial}, and TimeMoE~\cite{shi2025time},
MOIRAI~\cite{woo2024moirai}, and Chronos~\cite{ansari2024chronos} variants
retrained on BLAST. All downstream comparisons are zero-shot. We report mean
squared error (MSE) and mean absolute error (MAE),
using the triangularly weighted quantile combination in
Equation~\ref{eq:point-forecast} as the FlexCast point forecast.

\noindent
\textbf{Implementation details.}
Table~\ref{tab:experimental_settings} summarizes the default model, masking,
and optimization settings. Unless otherwise specified, FlexCast uses
single-pass full-horizon forecasting. We train FlexCast on eight NVIDIA A100
80GB GPUs with 800GB system RAM and two Intel Xeon Platinum 8358P CPUs at
2.60 GHz (64 CPU cores total). The software environment uses PyTorch 2.5.1
with CUDA 12.4.

\begin{table}[tbp]
  \caption{Common experimental settings used in the ablation studies.}
  \label{tab:experimental_settings}
  \centering
  \small
  \newcommand{\catlabel}[1]{\makebox[0.6cm][c]{\rotatebox[origin=c]{90}{#1}}}
  \begin{tabular}{@{}c@{\hspace{4pt}}ll@{}}
    \toprule
     & Setting & Value \\
    \midrule
    \multirow{7}{*}{\catlabel{Model Architecture}} & Maximum context window & 4096 \\
    & Patch length & 16 \\
    & Quantile levels & 99 \\
    & Encoder layers & 10 \\
    & Model dimension & 1024 \\
    & Attention heads & 16 \\
    & Feed-forward dimension & 4096 \\
    \midrule
    \catlabel{Mask} & Default mask-rate range & $[0.1, 0.3]$ \\
    \midrule
    \multirow{11}{*}{\catlabel{Training Optimization}} & Optimizer & AdamW \\
    & Effective batch size & 2048 \\
    & Training steps & 10k \\
    & Warmup ratio & 10\% \\
    & Learning-rate schedule & Cosine with linear warmup \\
    & Maximum learning rate & $3 \times 10^{-4}$ \\
    & Minimum learning rate & $1 \times 10^{-5}$ \\
    & Weight decay & 0.1 \\
    & AdamW $(\beta_1,\beta_2)$ & $(0.9, 0.95)$ \\
    & Gradient norm clipping & 1 \\
    & Precision & BF16 \\
    \bottomrule
  \end{tabular}
\end{table}

The 4096-step context window in Table~\ref{tab:experimental_settings} denotes
the maximum sequence extent processed during pretraining rather than a fixed
budget of observed history. For each training instance, this extent is
partitioned into visible observations, a randomly positioned contiguous
prediction span, and, when necessary, padding according to the effective
sequence length. Sampling the mask rate as $r\sim U(0.1,0.3)$ therefore
exposes the model to a broad distribution of prediction lengths and visible
context configurations. We adopt this interval because it achieves the best
performance among the evaluated mask-rate ranges.

\subsection{Evaluation-time Context and Latency Protocol}

For the main zero-shot comparison, we fix the observed context length of
FlexCast and PatchTST-FM to 3072, which is the longest context length used in
the BLAST evaluation. For Sundial, we use a sequence length of 2880, which is
the maximum context length supported by its reported setting. This protocol
separates the common evaluation-time history budget from each model's
pretraining window, ensuring that FlexCast receives no additional observed
history while evaluating its ability to compose context and horizon lengths
through the learned mask-and-predict interface. FlexCast uses single-pass
full-horizon inference in this comparison, directly predicting the whole
horizon in one model call.

For the latency study, we compare inference-time scaling on a single NVIDIA
A100 GPU with batch size 1, one input channel, a fixed context length of 512,
and prediction lengths of 96, 192, 336, and 720. We use 512 observations as the
common history because a 3376-step context is not uniformly supported by the
public implementations of all compared models. All evaluated implementations
can process the common 512-step context and generate the four horizons, while
all other parameters follow each model's default configuration. The 720-step
endpoints are measured; shorter-horizon baseline points are extrapolated for the
scaling comparison. The three-step FlexCast trend is estimated from the
corresponding one-step measurements. FlexCast
predicts the whole horizon in one forward pass, while next-token TSFMs generate
the target autoregressively and must repeat decoding steps. Consequently,
FlexCast latency grows only moderately with prediction length, whereas
sequential models such as Sundial, TimeMoE, and Chronos incur increasing
decoding costs over longer horizons. This advantage primarily comes from
avoiding repeated model calls and, relative to the larger baseline variants,
from a smaller parameter count.

Single-pass FlexCast achieves zero-shot accuracy competitive with or
better than strong TSFM baselines while being substantially faster than
sequential TSFMs. The accuracy advantage is most pronounced against models
trained on the same BLAST corpus, confirming that the flexible mask-and-predict
objective contributes beyond what the data alone provides. The latency
advantage over sequential forecasters follows from single-pass generation,
while the gap to other single-pass models is associated with FlexCast's compact
fixed-patch, direct-quantile inference path.

\section{Additional Experimental Results}\label{app:additional-results}

\subsection{Model Scaling Analysis}\label{app:exp-scaling-results}

Table~\ref{tab:scaling_law} provides the detailed model configurations and
average MSE values supporting the analysis in Section~\ref{sec:exp-scaling}.

\begin{table}[H]
  \caption{Model capacity analysis. Average MSE across six benchmarks (four horizons each) for three FlexCast model sizes. All variants use the same BLAST pretraining data, training steps, and default mask-rate range.}
  \label{tab:scaling_law}
  \centering
  \small
  \begin{tabular}{@{}lccccc@{}}
    \toprule
    Model & \#Params & Layers & $d_{\text{model}}$ & $d_{\text{ff}}$ & Avg MSE \\
    \midrule
    FlexCast$_{s}$ & 15M  & 6  & 256  & 1024 & .360 \\
    FlexCast$_{b}$ & 40M  & 8  & 512  & 2048 & .332 \\
    FlexCast$_{l}$ & 131.4M & 10 & 1024 & 4096 & \textbf{.308} \\
    \bottomrule
  \end{tabular}
\end{table}

\subsection{Model Parameter Sensitivity Analysis}\label{app:exp-sensitivity-results}

\noindent
\textbf{Context-length generalization.}
We first evaluate whether FlexCast can use different amounts of historical
context without retraining. The same checkpoint is evaluated with input
sequence lengths from 512 to 4096 and a fixed prediction horizon of 336. This
setting tests whether the flexible-context objective learned during pretraining
transfers to shorter and longer observed contexts at inference time.

\begin{figure}[tbp]
  \centering
  \includegraphics[width=\columnwidth]{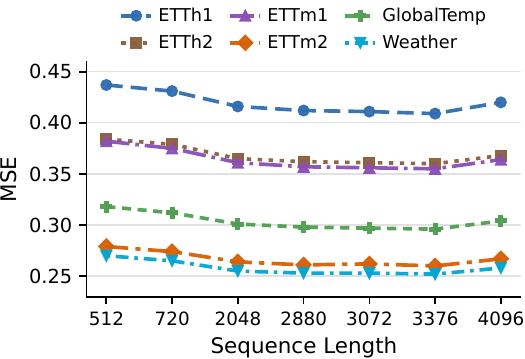}
  \caption{Context-length sensitivity across datasets with prediction length 336.}
  \Description{Line chart showing how forecasting MSE changes as the input context length varies across datasets.}
  \label{fig:context_horizon_sensitivity}
\end{figure}

Across datasets, Figure~\ref{fig:context_horizon_sensitivity} shows a consistent
context-length pattern. Short contexts, such as 512 or 720 steps, give the
highest MSE because they provide limited historical evidence. Increasing the
context to 2048, 2880, and 3072 steps steadily improves accuracy, and the best
results usually appear around 3376 steps. However, using the full 4096-step
context does not further improve performance and slightly increases MSE on all
datasets. This pattern reflects the training design: BLAST pretraining
sequences have length 4096, and FlexCast learns flexible context patterns within
this window.

\noindent
\textbf{Architecture sensitivity.}
We study whether FlexCast depends on a narrow set of hyperparameter choices or
remains stable across controlled changes. We use the default configuration and
create variants by changing one factor at a time while keeping other settings
aligned with Table~\ref{tab:experimental_settings}. All configurations use
BLAST pretraining, the quantile pinball objective, full-sequence inputs, 10k
training steps, batch size 256 per worker, and learning rate
$3\times10^{-4}$. For attention-head count, we evaluate
$n_{\mathrm{heads}}\in\{8,16,32,64\}$; for encoder depth,
$e_{\mathrm{layers}}\in\{6,10,12,15\}$; and for model dimension,
$d_{\mathrm{model}}\in\{256,512,1024,1536\}$, with the feed-forward dimension
scaled accordingly.

\begin{figure}[H]
  \centering
  \includegraphics[width=0.94\columnwidth]{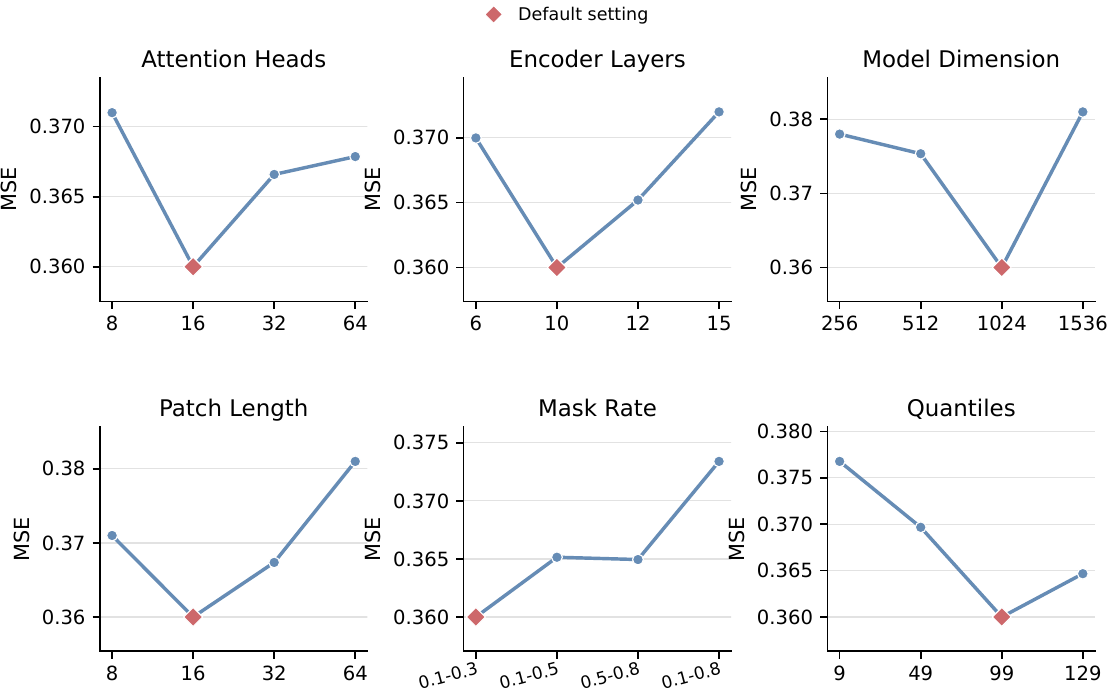}
  \caption{Parameter sensitivity on ETTh2 (context 3376, horizon 336). Each subplot varies one hyperparameter and marks the default.}
  \Description{Three sensitivity plots showing MSE changes under different attention head counts, encoder depths, and model dimensions.}
  \label{fig:parameter_sensitivity_results}
\end{figure}

Figure~\ref{fig:parameter_sensitivity_results} shows that the default
configuration has the lowest MSE among the tested settings. For attention
heads, 16 heads performs best, while 8, 32, and 64 heads have higher MSE. For
encoder depth, 10 layers performs best and 12 layers is close; neither the
shallower nor deeper variants improve on 10 layers under the fixed training
budget. For model dimension, 1024 obtains lower MSE than the smaller widths and
the 1536-dimensional variant, so increasing width beyond 1024 does not improve
accuracy under the fixed data and optimization budget.

\noindent
\textbf{Objective sensitivity.}
We further search over three objective-side parameters. For temporal
tokenization, we evaluate patch lengths of 8, 16, 32, and 64. For the masking
objective, we evaluate mask-rate intervals $[0.1,0.3]$, $[0.1,0.5]$,
$[0.5,0.8]$, and $[0.1,0.8]$. For probabilistic prediction, we evaluate 9, 49,
99, and 129 output quantiles.

A patch length of 16 performs best: shorter patches increase the token sequence
length and optimization burden, while longer patches lose temporal resolution.
The default mask interval $[0.1,0.3]$ is also the best among the tested choices.
Moderate masking exposes FlexCast to varied local and medium-range prediction
tasks without overwhelming the model with excessively long or ambiguous
targets within the fixed training budget. Finally, 99 quantiles outperform 9
and 49 quantiles, while increasing to 129 quantiles does not provide a further
gain.

\noindent
\textbf{Discussion.}
FlexCast is stable around the default design across both architecture and
objective parameters. The sensitivity results suggest two practical guidelines:
(1)~moderate model capacity (10 layers, $d{=}1024$, 16 heads) is sufficient for
the BLAST-scale pretraining budget, and further scaling without additional data
or training steps yields diminishing returns; (2)~the mask-rate interval should
be kept moderate ($[0.1,0.3]$) to produce diverse but learnable context
configurations, as overly aggressive masking creates targets that exceed what
the model can reliably cover within the training budget.

\subsection{Expanded Discussion of Core Ablations}

The flexible-context study clarifies why context diversity is the
stronger factor. A model trained with a single context length can specialize to
one conditioning distribution, while foundation-model evaluation often
presents different context budgets. Sampling multiple effective contexts
reduces this mismatch and encourages forecasting rules that remain stable under
different amounts of historical evidence. Flexible targets are still useful,
but their role is complementary: they teach the model to handle different
prediction granularities once the conditioning distribution is sufficiently
broad.

The masking ablation shows the same principle from the target side. Tail
masking matches the downstream forecasting layout but creates a narrow
pretraining distribution. Scattered short masks provide more positions, yet
they make the task closer to local interpolation and reduce the pressure to
model long-range temporal structure. Random-rate contiguous masking occupies
the middle ground: each target remains a coherent forecasting-like segment,
while its position and length vary enough to create many context-target
patterns within one model. Thus, FlexCast preserves the useful learning-side
diversity of AR training without requiring sequential rollout at inference
time.

Confidence-based multi-step inference remains close to one-step inference in
most settings, with changes typically within a few percent. Its clearest
improvement appears on ETTh2 at the 720-step horizon, where relative MSE
decreases by 4.7\%. This suggests that confidence-ordered generation can help
when the horizon is long and uncertainty is high, without requiring
left-to-right generation. It retains the mask-and-predict interface and changes
only which target patches are revealed across rounds, following the same
principle as MaskGIT-style generation~\cite{chang2022maskgit}.

Blockwise AR inference instead uses a fixed left-to-right reveal order.
A 192-step prediction unit is generally worse than a 96-step unit, showing that
autoregressive decoding is sensitive to prediction granularity. Even with the
shorter unit, blockwise inference usually increases MSE relative to one-step
inference because generated values are appended to the context and early errors
condition later predictions. Single-pass inference is therefore sufficient in
most settings; when additional rounds are useful, confidence ordering provides
a better multi-step alternative without enforcing a fixed generation order.

\balance
\section{Limitations and Future Work}\label{sec:limitations_future_work}

In the evaluated BLAST setting, FlexCast recovers much of the benefit associated
with autoregressive rollout through flexible context learning and efficient
single-pass inference. The implementation has three limitations.

\noindent
\textbf{Training Efficiency under Variable Lengths.}
Although FlexCast supports variable effective lengths through padding masks, the
current batching strategy is not fully compute efficient. Training instances
with different effective lengths are padded to a common maximum length in the
batch, and padding patches are excluded from attention and loss. This prevents
the model from learning padding artifacts, but it does not remove all wasted
computation because padded positions still occupy tensor slots and determine the
sequence shape processed by the Transformer. The inefficiency becomes more
visible when the pretraining corpus contains many short sequences or when the
sampled context configurations have highly diverse lengths.
Packed variable-length training could reduce this waste. Following Patch n'
Pack~\cite{dehghani2023patch},
multiple independent short samples could be concatenated into the same
Transformer row until the token budget is filled. A block-structured attention
mask would prevent tokens from attending across sample boundaries, while each
packed sample would retain its own context, prediction, and padding masks. This
would preserve the FlexCast objective while improving token utilization, so that
the same training budget covers more valid patches and more diverse
context configurations.

\noindent
\textbf{Adaptive Inference Schedules.}
Our experiments mainly compare single-pass forecasting, confidence-ordered
iterative generation, and left-to-right blockwise AR inference under a fixed
pretraining protocol. More adaptive reveal schedules may further improve
long-horizon or high-uncertainty forecasts, for example by changing the number
of generation rounds according to the predicted quantile spread. Such schedules
could keep the single-pass path as the default for easy cases while allocating
additional model calls only to horizons or patches with high predictive
uncertainty.

\noindent
\textbf{Scaling to Broader Pretraining Regimes.}
The current study focuses on the BLAST-style pretraining and evaluation setting
and does not test larger, more heterogeneous corpora. In particular, the
optimal mask-rate distribution may
change with model size, data mixture, and token budget: larger models may handle
broader and harder context configurations, while smaller models may benefit
from a narrower curriculum.

\end{document}